# Survey Propagation Revisited

**Lukas Kroc** **Ashish Sabharwal** **Bart Selman**

Department of Computer Science, Cornell University, Ithaca, NY 14853-7501, U.S.A.*
{kroc,sabhar,selman}@cs.cornell.edu

## Abstract

Survey propagation (SP) is an exciting new technique that has been remarkably successful at solving very large hard combinatorial problems, such as determining the satisfiability of Boolean formulas. In a promising attempt at understanding the success of SP, it was recently shown that SP can be viewed as a form of belief propagation, computing marginal probabilities over certain objects called covers of a formula. This explanation was, however, shortly dismissed by experiments suggesting that non-trivial covers simply do not exist for large formulas. In this paper, we show that these experiments were misleading: not only do covers exist for large hard random formulas, SP is surprisingly accurate at computing marginals over these covers despite the existence of many cycles in the formulas. This re-opens a potentially simpler line of reasoning for understanding SP, in contrast to some alternative lines of explanation that have been proposed assuming covers do not exist.

## 1 INTRODUCTION

Survey Propagation (SP) is a new exciting algorithm for solving hard combinatorial problems. It was discovered by Mezard, Parisi, and Zecchina (2002), and is so far the only known method successful at solving random Boolean satisfiability (SAT) problems with 1 million variables and beyond in near-linear time in the hardest region. The SP method is quite radical in that it tries to approximate certain marginal probabilities related to the set of satisfying assignments. It then iteratively assigns values to variables with the most extreme probabilities. In effect, the algorithm behaves like the usual backtrack search methods for SAT (DPLL-based), which also assign variable values incrementally in an attempt to find a satisfying assignment. However, quite surprisingly, SP almost never has to backtrack. In other words, the "heuristic guidance" from SP is almost always correct. Note that, interestingly, computing marginals on satisfying assignments is actually believed to be much harder than finding a single satisfying assignment (#P-complete vs. NP-complete). Nonetheless, SP is able to efficiently approximate certain marginals and uses this information to successfully find a satisfying assignment.

SP was derived from rather complex statistical physics methods, specifically, the so-called cavity method developed for the study of spin glasses. Close connections to belief propagation (BP) methods were subsequently discovered. In particular, it was discovered by Braunstein and Zecchina (2004) (later extended by Maneva, Mossel, and Wainwright (2005)) that SP equations are equivalent to BP equations for obtaining marginals over a special class of combinatorial objects, called covers. Intuitively, a cover provides a representative generalization of a cluster of satisfying assignments. The discovery of a close connection between SP and BP via the use of covers laid an exciting foundation for explaining the success of SP. Unfortunately, subsequent experimental evidence suggested that hard random 3-SAT formulas have, with high probability, only one (trivial) cover (Maneva et al., 2005). This would leave all variables effectively in an undecided state, and would mean that marginals on covers cannot provide any useful information on how to set variables. Since SP clearly sets variables in a non-trivial manner, it was conjectured that there must be another explanation for the good behavior of SP; in particular, one that is not based on the use of marginal probabilities of variables in the covers.

In this paper, we revisit the claim that hard random 3-SAT formulas do not have interesting non-trivial cov-

*Research supported by Intelligent Info. Systems Instt. (IISI), Cornell Univ., AFOSR grant FA9550-04-1-0151.

ers. In fact, we show that such formulas have large numbers of non-trivial covers. The main contribution of the paper is the first clear empirical evidence showing that in random 3-SAT problems near the satisfiability and hardness threshold, (1) a significant number of non-trivial covers exist; (2) SP is remarkably good at computing variable marginals based on covers; and (3) these cover marginals closely relate to solution marginals at least in the extreme values, where it matters the most for survey inspired decimation. As a consequence, we strongly suspect that explaining SP in terms of covers may be the correct path after all.

Note that (2) above is quite surprising for random 3-SAT formulas because such formulas have many loops. The known formal proof that SP computes cover marginals only applies to tree-structured formulas, which in fact have only a single (trivial) cover. Further, it's amazing that while SP computes such marginals in a fraction of a second, the next best methods of computing these marginals that we know of (via exact enumeration, or sampling followed by "peeling") require over 100 CPU hours.

Our experiments also indicate that cover marginals are more "conservative" than solution marginals in the sense that variables that are extreme with respect to cover marginals are almost certainly also extreme with respect to solution marginals, but not vice versa. This sheds light on why it is safe to set variables with extreme cover marginals in an iterative manner, as is done in the survey inspired decimation process for finding a solution using the marginals computed by SP.

In addition to these empirical results, we also revisit the derivation of the SP equations themselves, with the goal of presenting the derivation in an insightful form purely within the realm of combinatorial constraint satisfaction problems (CSPs). We describe how one can reformulate in a natural step-by-step manner the problem of finding a satisfying assignment into one of finding a cover, by considering related factor graphs on larger state spaces. The BP equations for this reformulated problem are exactly the SP equations for the original problem, as shown in the Appendix.

## 2 COVERS OF CNF FORMULAS

We start by introducing the notation and the basic concepts that we use throughout the paper. We are concerned with Boolean formulas in Conjunctive Normal Form or CNF, that is, formulas of the form $F \equiv (l_{11} \vee \ldots \vee l_{1k_1}) \wedge \ldots \wedge (l_{m1} \vee \ldots \vee l_{mk_m})$, where each $l_{ik}$ (called a *literal*) is a Boolean variable $x_j$ or its negation $\neg x_j$. Each conjunct of $F$, which itself is a disjunction of literals, is called a *clause*. In 3-CNF or 3-SAT formulas, every clause has 3 literals. Random 3-SAT formulas over $n$ variables are generated by uniformly randomly choosing a pre-specified number of clauses over these $n$ variables. The Boolean satisfiability problem is the following: Given a CNF formula $F$ over $n$ variables, find a truth assignment $\sigma$ for the variables such that every clause in $F$ evaluates to TRUE; $\sigma$ is called a *satisfying assignment* or a *solution* of $F$. We identify TRUE with 1 and FALSE with 0.

A truth assignment to $n$ variables can be viewed as a string of length $n$ over the alphabet $\{0, 1\}$, and extending this alphabet to include a third letter "$*$" leads to a *generalized assignment*. A variable with the value $*$ can be interpreted as being "undecided," while variables with values 0 or 1 can be interpreted as being "decided" on what they want to be. We will be interested in certain generalized assignments called *covers*. Our formal definition of covers follows the one given by Achlioptas and Ricci-Tersenghi (2006). Let variable $x$ be called a *supported variable* under a generalized assignment $\sigma$ if there is a clause $C$ such that $x$ is the only variable that satisfies $C$ and all other literals of $C$ are FALSE. Otherwise, $x$ is called *unsupported*.

**Definition 1.** A generalized assignment $\sigma \in \{0, 1, *\}^n$ is a **cover** of a CNF formula $F$ iff

1. every clause of $F$ has at least one satisfying literal or at least two literals with value $*$ under $\sigma$, and
2. $\sigma$ has no unsupported variables assigned 0 or 1.

The first condition ensures that each clause of $F$ is either already satisfied by $\sigma$ or has enough undecided variables to not cause any undecided variable to be forced to decide on a value (no "unit propagation"). The second condition says that each variable that is assigned 0 or 1 is set that way for a reason: there exists a clause that relies on this setting in order to be satisfied. For example, consider the formula $F \equiv (x \vee \neg y \vee \neg z) \wedge (\neg x \vee y \vee \neg z) \wedge (\neg x \vee \neg y \vee z)$. $F$ has exactly two covers: 111 and $***$. This can be verified by observing that whenever some variable is 0 or $*$, then all non-$*$ variables are unsupported. Notice that the string of all $*$'s always satisfies the conditions in Definition 1; we refer to this string as the *trivial cover*.

Covers were introduced by Maneva et al. (2005) as a useful concept to analyze the behavior of SP, but their combinatorial properties are much less known than those of solutions. A cover can be thought of as a *partial assignment* to variables, where the variables assigned $*$ are considered unspecified. In this sense, each cover is a representative of a potentially large set of complete truth assignments, satisfying as well as not satisfying. This motivates further differentiation:

**Definition 2.** A cover $\sigma \in \{0, 1, *\}^n$ of $F$ is a **true cover** iff there exists a satisfying assignment $\tau \in \{0, 1\}^n$ of $F$ such that $\sigma$ and $\tau$ agree on all values where

$\sigma$ is not a $*$, i.e., $\forall i \in \{1, \ldots, n\} (\sigma_i \neq * \implies \sigma_i = \tau_i)$. Otherwise, $\sigma$ is a **false cover**.

A true cover thus generalizes at least one satisfying assignment. True covers are interesting to study when trying to satisfy a formula, because if there exists a true cover with variable $x$ assigned 0 or 1, then there must also exist a satisfying assignment with the same setting of $x$.

One can construct a true cover $\sigma \in \{0, 1, *\}^n$ of $F$ by starting with any satisfying assignment $\tau \in \{0, 1\}^n$ of $F$ and generalizing it using a simple procedure called **$*$-propagation**.[1] The procedure starts by initially setting $\sigma = \tau$. It then repeatedly chooses an arbitrary variable unsupported under $\sigma$ and turns it into a $*$, until there are no more unsupported variables. The resulting string $\sigma$ is a true cover, which can be verified as follows. The satisfying assignment $\tau$ already satisfies the first condition in Definition 1, and $*$-propagation does not destroy this property. In particular, a variable on which some clause relies is never turned into a $*$. The second condition in Definition 1 is also clearly satisfied when $*$-propagation halts, so that $\sigma$ must be a cover. Moreover, since $\sigma$ generalizes $\tau$, it is a true cover. Note that $*$-propagation can, in principle, be applied to an arbitrary generalized assignment. However, unless we start with one that satisfies the first condition in the cover definition, $*$-propagation may not lead to a cover.

We end with a discussion of two insightful properties of covers. The first relates to "self-reducibility" and the second to covers for tree-structured formulas.

**No self-reducibility.** Consider the relation between the decision and search versions of the problem of finding a solution of a CNF formula $F$. In the decision version, one needs an algorithm that determines whether or not $F$ has a solution, while in the search version, one needs an algorithm that explicitly finds a solution. The problem of finding a solution for $F$ is *self-reducible*, i.e., given an oracle for the decision version, one can efficiently solve the search version by iteratively fixing variables to 1 or 0, testing whether there is still a solution, and continuing in this way. Somewhat surprisingly, this strategy does *not* work for the problem of finding a cover. In other words, an oracle for the decision version of this problem does not immediately provide an efficient algorithm for finding a cover. (The lack of self-reducibility makes it very hard to find covers as we will see below.) As a concrete example, consider the formula $F$ described right after Definition 1. To construct a cover of $F$, we could ask whether there exists a cover with $x$ set to 1. Since 111 is a cover (yet unknown to us), the decision oracle would say yes. We could then fix $x$ to 1, simplify the formula to $(y \vee \neg z) \wedge (\neg y \wedge z)$, and ask whether there is a cover with $y$ set to 0. This residual formula indeed has 00 as a cover, and the oracle would say yes. With one more query, we will end up with 100 as the values of $x, y, z$, which is in fact *not* a cover of $F$.

[1]This was introduced under different names as the peeling procedure or coarsening, e.g., by Maneva et al. (2005).

**Tree-structured formulas.** For tree-structured formulas without unit clauses, i.e., formulas whose factor graph does not have a cycle, the only cover is the trivial all-$*$ cover. We argue this using the connection between covers and SP shown by Braunstein and Zecchina (2004), which says that when generalized assignments have a uniform prior, SP on a tree formula $F$ provably computes probability marginals of variables being 0, 1, and $*$ in covers of $F$. Moreover, it can be verified from the iterative equations for SP that with no unit clauses, zero marginals for any variable being 0 or 1, and full marginals for any variable being a $*$ is a fixed point of SP. Since SP provably has exactly one fixed point on tree formulas, it follows that the only cover of such formulas is the trivial all-$*$ cover.

## 3 PROBLEM REFORMULATION: FROM SOLUTIONS TO COVERS

We now show that the concept of covers can be quite naturally arrived at when trying to find solutions of a CNF formula, thus motivating the study of covers from a purely generative perspective. Starting with a CNF formula $F$, we describe how $F$ is transformed step-by-step into the problem of finding covers of $F$, motivating each step.

Although our discussion applies to any CNF formula $F$, we will be using the following example formula with 3 variables and 4 clauses to illustrate the steps:

$$\underbrace{(x \vee y \vee \neg z)}_{a} \wedge \underbrace{(\neg x \vee y)}_{b} \wedge \underbrace{(\neg y \vee z)}_{c} \wedge \underbrace{(x \vee \neg z)}_{d}$$

Let $N$ denote the number of variables, $M$ the number of clauses, and $L$ the number of literals of $F$.

**Original problem.** The problem is to find an assignment in the space $\{0, 1\}^N$ that satisfies $F$. The *factor graph* for $F$ has $N$ variable nodes and $M$ function nodes, corresponding directly to the variables $x, y, \ldots$ and clauses $a, b, \ldots$ in $F$ (see e.g. Kschischang et al. (2001)). The factor graph for the example formula is depicted below. Here factors $F_a, F_b, \ldots$ represent predicates ensuring that the corresponding clause has at least one satisfying literal.

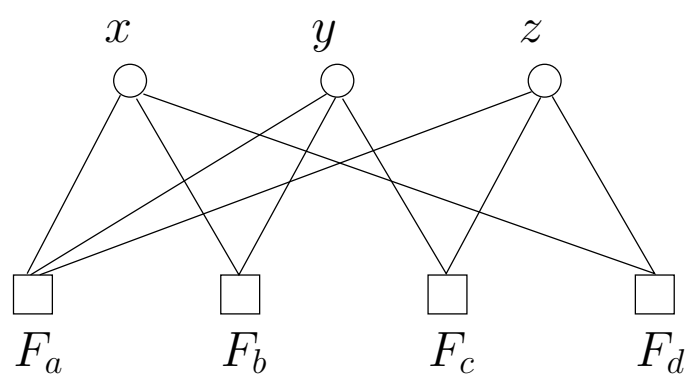

**Variable occurrences.** The first step in the transformation is to start treating every *variable occurrence* $x_a, x_b, y_a, y_b, \dots$ in $F$ as a separate unit that can be either 0 or 1. This allows for more flexibility in the process of finding a solution, since a variable can decide what value to assume in each clause separately. Of course, we need to add constraints to ensure that the occurrence values are eventually consistent: for every variable $x$ in $F$, we add a constraint $F_x$ that all occurrences of $x$ have the same value. Now the search space is $\{0,1\}^L$, and the corresponding factor graph contains $L$ variable nodes and $M+N$ function nodes (the original clause factors $F_a, F_b, \dots$ and the new constraints $F_x, F_y, \dots$).

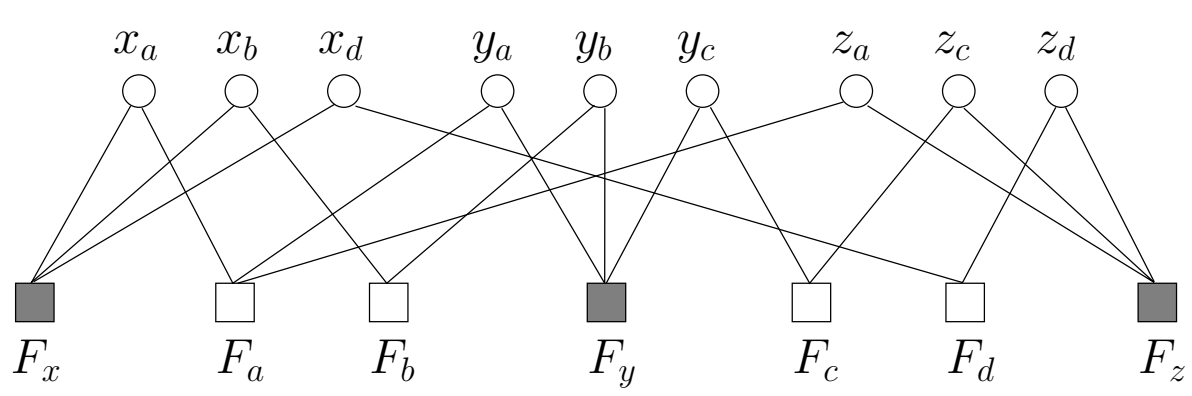

At this point, we have not relaxed solutions to the original problem $F$: solutions to the modified problem correspond precisely to the original solutions, because variable occurrences are forced to be consistent. However, we moved this consistency check from the syntactic level (variables could not be inconsistent simply by the problem definition) to the semantic level (we have special constraints to guarantee consistency).

**Relaxing assignments.** The next step is to relax the problem by allowing variable nodes to assume the special value "$*$". The semantics of $*$ is "undecided," meaning that the variable node is set neither to 0 nor to 1. The new search space is $\{0,1,*\}^L$, and we must specify how our constraints handle the value $*$. Variable constraints $F_x, \dots$ have the same meaning as before, namely, all variable nodes $x_a, x_b, \dots$ have the same value for every variable $x$. Clause constraints $F_a, \dots$ now have a modified meaning: a clause is satisfied if it contains at least one satisfying literal or at least *two* literals with the value $*$. The motivation here is to either satisfy a clause or leave enough "freedom" in the form of at least two undecided variables. (A single undecided variable would be forced to take on a particular value if all other literals in the clause were falsified.) With this transformation, the factor graph remains structurally the same, while the set of possible values for variable nodes changes.

The solutions to this modified problem do not necessarily correspond directly to solutions of the original one. In particular, if there are no unit clauses and all variables are set to $*$, the problem is already "solved" without providing any useful information.

**Reducing freedom of choice.** To distinguish variables that could assume the value $*$ from those that truly need to be fixed to either 0 or 1, we require that every non-$*$ variable has a clause that needs the variable to be 0 or 1 in order to be satisfied. The search space does not change, but we need to add constraints to implement the reduction in the freedom of choice.

Notice that this requirement is equivalent to "no unsupported variables" in the definition of a cover, and that the first requirement in that definition is fulfilled by the clause constraints. Therefore, we are now searching for covers of $F$. A natural way to represent the "no unsupported variable" constraint in the factor graph is to add for each variable $x$ a new function node $F'_x$, connected to the variable nodes for $x$ as well as for all other variables sharing a clause with $x$. This, of course, creates many new links and introduces additional short cycles, even if the original factor graph was acyclic. The following transformation step alleviates this issue.

**Reinterpreting variable nodes.** As the final step, we change the semantics of the variable nodes' values and of the constraints so that the "no unsupported variable" condition can be enforced without additional function nodes. The reasoning is that the simple $\{0,1,*\}$ domain creates a bottleneck for how much information can be communicated between nodes in the factor graph. By altering the semantics of the variable nodes' values, we can improve on this.

The new value of a variable node $x_a$ will be a pair $(r_{a\to x}, w_{x\to a}) \in \{(0,0),(0,1),(1,0)\}$, so that the size of the search space is still $3^L$. We interpret the value $r_{a\to x}$ as a *request* from clause $a$ to variable $x$ with the meaning that $a$ relies on $x$ to satisfy it, and the value $w_{x\to a}$ as a *warning* from variable $x$ to clause $a$ that $x$ is set such that it does not satisfy $a$. The values 1 and 0 indicate presence and absence, resp., of the request or warning. We can recover the original $\{0,1,*\}$ values from these new values as follows: if $r_{a\to x}=1$ for some $a$, then $x$ is set to satisfy clause $a$; if there is no request from any clause where $x$ appears, then $x$ is undecided (a value of $*$ in the previous interpretation). The variable constraints $F_x, \dots$ not only ensure consistency of the values of $x_a, x_b, \dots$ as before, but also ensure the second cover condition as described below. The clause constraints $F_a, \dots$ remain unchanged.

The variable constraint $F_x$ is a predicate ensuring that the following two conditions are met:

1. if $r_{a\to x} = 1$ for *any* clause $a$ where $x$ appears, then $w_{x\to b} = 0$ for all clauses $b$ where $x$ appears with the *same* sign as in $a$, and $w_{x\to b} = 1$ for all $b$ where $x$ appears with the *opposite* sign. Since $x$ must be set to satisfy $a$, this ensures that clauses that are unsatisfied by $x$ do receive a warning.
2. if $r_{a\to x} = 0$ for *all* clauses $a$ where $x$ appears, then $w_{x\to a} = 0$ for all of them, i.e., no clause receives a warning from $x$.

To evaluate $F_x$, values $(r_{a\to x}, w_{x\to a})$ are needed only for clauses $a$ in which $x$ appears, which is exactly the set of variable nodes the factor $F_x$ is connected to. Notice that the case $(r_{a\to x}, w_{x\to a}) = (1, 1)$ cannot happen due to condition 1 above. The conditions also imply that the variable occurrences of $x$ are consistent, and in particular that two clauses where $x$ appears with opposite signs (say $a$ and $b$) cannot simultaneously request to be satisfied by $x$. This is because either $r_{a\to x} = 0$ or $r_{b\to x} = 0$ must hold due to condition 1.

The clause constraint $F_a$ is a predicate stating that clause $a$ issues a request to its variable $x$ if and only if it receives warnings from all its other variables: $r_{a\to x} = 1$ iff $w_{y\to a} = 1$ for all variables $y \neq x$ in $a$. Again, $F_a$ can be evaluated using exactly values from the variable nodes it is connected to.

When clause $a$ issues a request to variable $x$ (i.e., $r_{a\to x} = 1$), $x$ must be set to satisfy $a$, thus providing a satisfying literal for $a$. If $a$ does not issue any request, then according to the condition of $F_a$, at least two of $a$'s variables, say $x$ and $y$, must not have sent a warning. In this case, $F_x$ and $F_y$ state that each of $x$ and $y$ is either undecided or satisfies $a$. Thus the first condition in the cover definition holds in any solution of this new constraint satisfaction problem. The second condition also holds, because every variable $x$ that is not undecided must have received a request from some clause $a$, so that $x$ is the only literal in $a$ that is not FALSE. Therefore $x$ is supported.

Let us denote this final constraint satisfaction problem by $P(F)$. (It is a function of the original formula $F$.) Notice that the factor graph of $P(F)$ has the same topology as the factor graph of $F$. In particular, if $F$ has a tree factor graph, so does $P(F)$. Further, by the construction of $P(F)$ described above, its solutions correspond precisely to the covers of $F$.

### 3.1 INFERENCE OVER COVERS

This section discusses an approach for solving the problem $P(F)$ with probabilistic inference using belief propagation (BP). It arrives at the survey propagation equations for $F$ by applying BP equations to $P(F)$.

Since the factor graph of $P(F)$ can be easily viewed as a Bayesian Network (cf. Pearl, 1988), one can compute marginal probabilities over the set of satisfying assignments of the problem, defined as

$$\Pr[x_a = v \mid \text{all constraints of } P(F) \text{ are satisfied}]$$

for each variable node $x_a$ and $v \in \{(0,0),(0,1),(1,0)\}$. The probability space here is over all assignments to variable nodes with uniform prior.

Once these solution marginals are known, we know which variables are most likely to assume a particular value, and setting these variables simplifies the problem. A new set of marginals can be computed on this simplified formula, and the whole process repeated. This method of searching for a satisfying assignment is called the **decimation procedure**. The problem, of course, is to compute the marginals (which, in general, is much harder than finding a satisfying assignment). One possibility for computing marginals is to use the belief propagation algorithm (cf. Pearl, 1988). Although provably correct essentially only for formulas with a tree factor graph, BP provides a good approximation of the true marginals in many problem domains in practice (Murphy et al., 1999). Moreover, as shown by Maneva et al. (2005), applying the BP algorithm to the problem of searching for covers of $F$ results in the SP algorithm. Thus, on formulas with a tree factor graph, the SP algorithm provably computes marginal probabilities over covers of $F$, which are equivalent to marginals over satisfying assignments of $P(F)$. When the formula contains loops, SP computes a loopy approximation to the cover marginals. Specific details of the derivation of SP equations from the problem $P(F)$ are deferred to the Appendix.

## 4 EXPERIMENTAL RESULTS

This section presents our main contributions. We begin by demonstrating that non-trivial covers do exist in large numbers in random 3-SAT formula, and then explore connections between SP, BP, and variable marginals computed from covers as well as solutions, showing in particular that SP approximates cover marginals surprisingly well.

### 4.1 EXISTENCE OF COVERS

Motivated by theoretical results connecting SP to covers of formulas, Maneva et al. (2005) suggested an experimental study to test whether non-trivial covers even exist in random 3-SAT formulas. They proposed a seemingly good way to do this (the "peeling experiment"), namely, start with a uniformly random satisfying assignment of a formula $F$ and, while it has unsupported variables, $*$-propagate the assignment. When the process terminates, one obtains a (true) cover of $F$. Unfortunately, what they observed is that this process repeatedly hits the trivial all-$*$ cover, from which they concluded that non-trivial covers most likely do not exist for such formulas. However, it is known that near-uniformly sampling solutions of such formulas to start with is a hard problem in itself and that most sampling methods obtain solutions in a highly non-uniform manner (Wei et al., 2004). Consequently, one must be careful in drawing conclusions from relatively few and possibly biased samples.

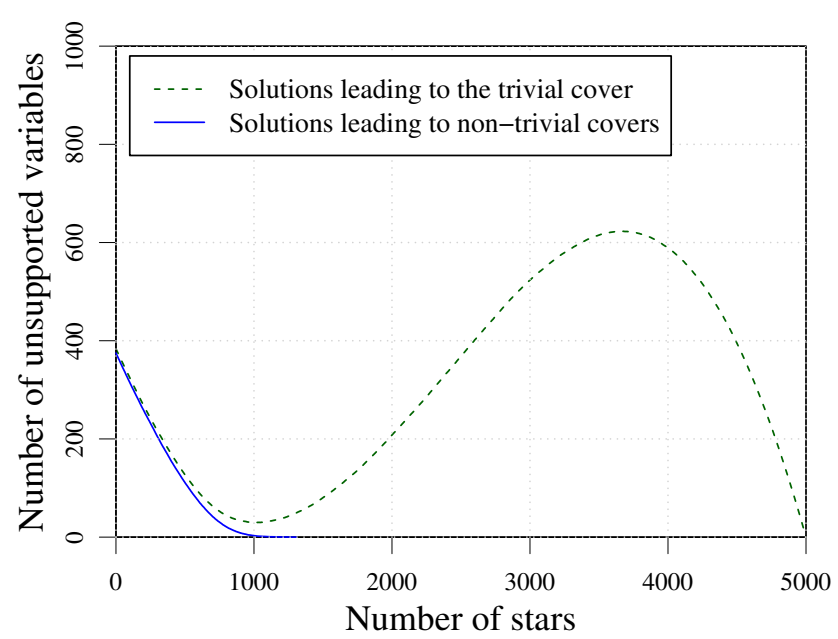

Figure 1: The peeling experiment, showing the evolution of the number of stars as $*$-propagation is performed.

To understand this issue better, we ran the same peeling experiment on a 5000 variable random 3-SAT formula at clause-to-variable ratio 4.2 (which is close to the hardness threshold for random 3-SAT problems), but used `SampleSat` (Wei et al., 2004) to obtain samples, which is expected to produce fairly uniform samples. Figure 1 shows the evolution of the number of unsupported variables at each stage as $*$-propagation is performed starting from a solution. Here, the x-axis shows the number of stars, which monotonically increases by $*$-propagation. The y-axis shows the number of unsupported variables present at each stage. As one moves from left to right following the $*$-propagation process, one hits a cover if the number of unsupported variables drops to zero (so that $*$-propagation terminates). The two curves in the plot correspond to solutions that $*$-propagated to the trivial cover and those that did not. In our experiment, out of 500 satisfying assignments used, nearly 74% led to the trivial cover; their average is represented by the top curve. The remaining 26% of the sampled solutions actually led to non-trivial covers; their average is represented by the bottom curve. Thus, when solutions are sampled near-uniformly, a substantial fraction of them lead to non-trivial covers.[2]

[2] That this was not observed by Maneva et al. (2005) can be attributed to the fact that SP was used to find satisfying assignments (Mossel, 2007), resulting in highly non-uniform samples.

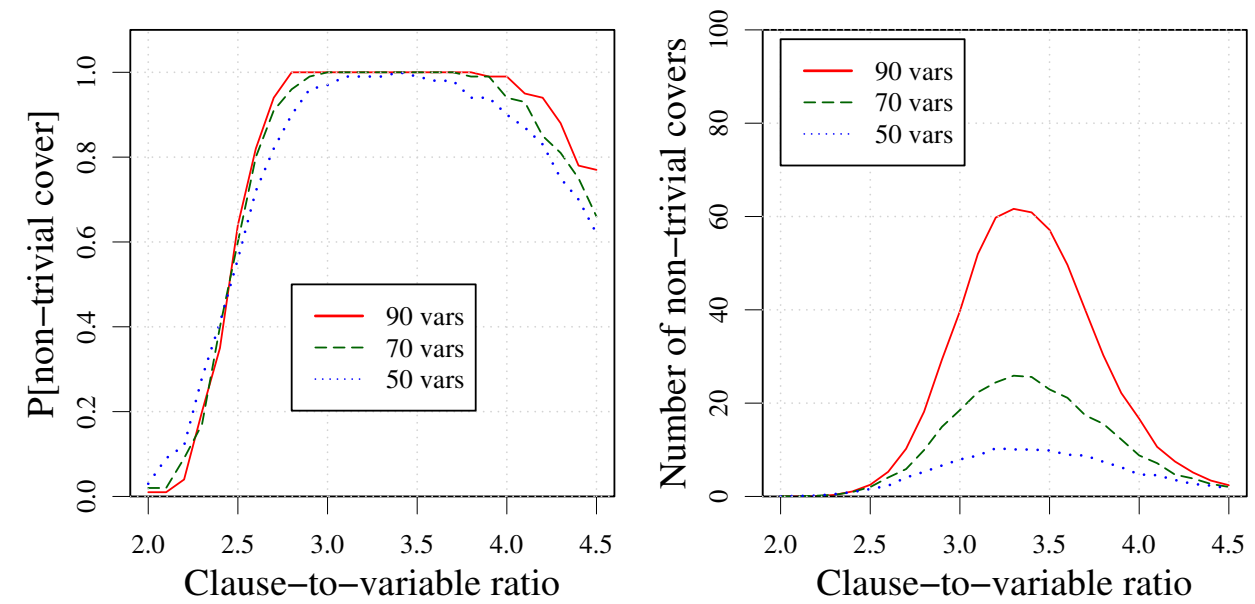

Figure 2: Non-trivial covers in random formulas. Left: existence probability. Right: average number.

An alternative method of finding covers is to create a new Boolean formula $G$ whose solutions correspond go the covers of $F$. It turned out to be extremely hard to solve $G$ to find *any* non-trivial cover using state-of-the-art SAT solvers for number of variables as low as 150. So we confined our experiments to small formulas, with 50, 70 and 90 variables. We found all covers for such formulas with varying clause-to-variable ratios $\alpha$. The results are shown in Figure 2, where each data point corresponds to statistics obtained from 500 formulas. The left pane shows the probability that a random formula, for a given clause-to-variable ratio, has at least one non-trivial cover (either true or false). The figure shows a nice phase transition where covers appear, at around $\alpha = 2.5$, which is surprisingly sharp given the small formula sizes. Also, the region where covers surely exist is widening on both sides as the number of variables increases, supporting the claim that non-trivial covers exist even in large formulas. The right pane of Figure 2 shows the actual number of non-trivial covers, with a clear trend that the number increases with the size of the formula, for all values of the clause-to-variable ratio. It is worth noting that the number of covers is very small compared to the number of satisfying assignments; e.g. for 90 variables and $\alpha = 4.2$, the expected number of satisfying assignments is $150,000$, while there are only 8 covers on average. Somewhat surprisingly, the number of false covers is almost negligible, around 2 at the peak, and does not seem to be growing nearly as fast as the total number of covers. This might explain why SP, although approximating marginals over all covers, is successful in finding satisfying assignments.

We also consider how the number of solutions that lead to non-trivial covers changes for larger formulas, as the number of variables $N$ increases from 200 to 4000. The left pane of Figure 3 shows that this number, in fact, *grows exponentially* with $N$. The number is computed by averaging over 20,000 sampled solutions from 200 formulas at ratio 4.2 for each $N$, estimating the fraction $p(N)$ of these that lead to a non-trivial cover, and scaling it up by the expected number of solutions, which is $(2 \times (7/8)^{4.2})^N \approx 1.1414^N$. (The number of solutions for such formulas at ratio 4.2 is known to be highly concentrated around its expectation.) The resulting number, $p(N) \times 1.1414^N$, is plotted on the y-axis of the left pane, with $N$ on the x-axis. The right pane of Figure 3 shows the data used to estimate the fraction $p(N)$ along with its fit on the y-axis, with $N$ on the x-axis again.

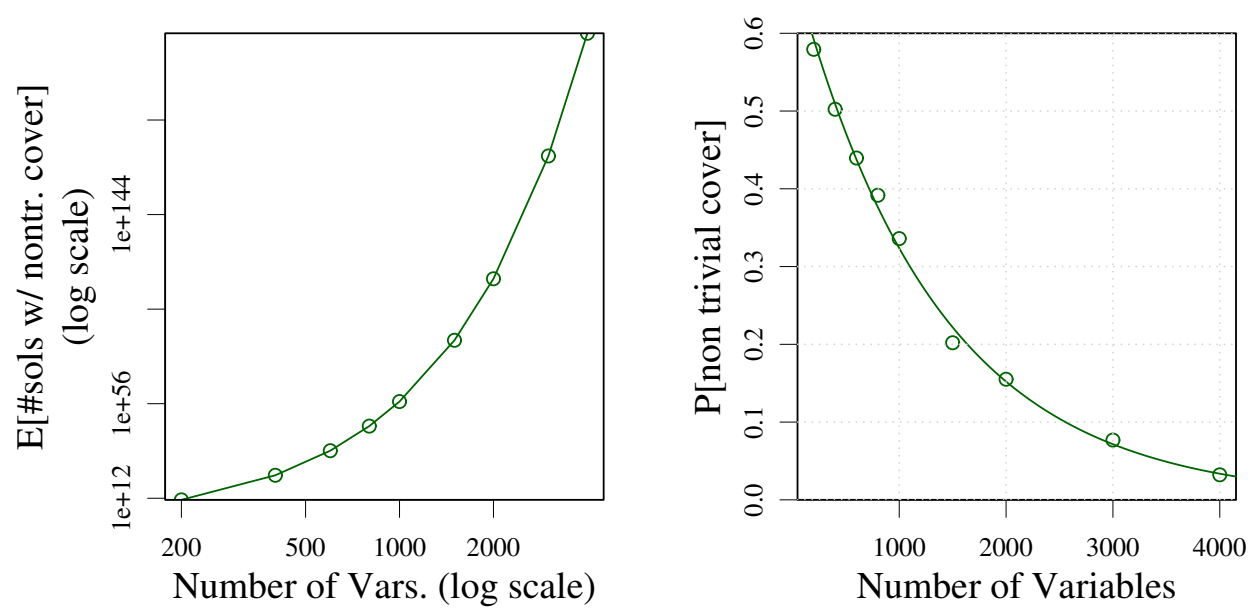

Figure 3: Left: Expected number of solutions leading to non-trivial covers (log-log scale). Right: Probability of a solution leading to a non-trivial cover.

Notice that the left pane is in log-scale for both axes, and clearly increases faster than a linear function. This shows that the expected number of solutions that lead to non-trivial covers grows super-polynomially. In fact, performing a best fit for this curve suggests that this number grows exponentially, roughly as $1.1407^N$. This number is indeed a vanishingly small fraction of the expected number of solutions ($1.1414^N$) as observed by Maneva et al. (2005), but nonetheless exponentially increasing. The existence of covers for random 3-SAT also aligns with what Achlioptas and Ricci-Tersenghi (2006) recently proved for $k$-SAT with $k \geq 9$.

### 4.2 SP, BP, AND MARGINALS

We now study the behavior of SP and BP on a random formula in relation to solutions and covers of that formula. While theoretical work has shown that SP, viewed as BP on a related combinatorial problem, provably computes cover marginals on tree-structured formulas, we demonstrate that even on random 3-SAT instances, which are far from tree-like, SP approximates cover marginals surprisingly well. We also show that cover marginals, especially in the extreme range, are closely related to solution marginals in an intriguing "conservative" fashion. *The combination of these two effects, we believe, plays a crucial role in the success of SP.* Our experiments also reveal that BP performs poorly at computing any marginals of interest.

Given marginal probabilities, we define the **magnetization** of a variable to be the difference between the marginals of the variable being positive and it being negative. For the rest of our experiments, we start with a random 3-SAT formula $F$ with 5000 variables and 21000 clauses (clause-to-variable ratio of 4.2), and plot the magnetization of the variables of $F$ in the range $[-1, +1]$.[3] The marginals for magnetization are obtained from four different sources, which are compared and contrasted against each other: (1) by running SP on $F$ till the iterations converge; (2) by running BP on $F$ but terminating it after 10,000 iterations because the equations do not converge; (3) by sampling solutions of $F$ using `SampleSat` and computing an estimate of the positive and negative marginals from the sampled solutions (the solution marginals); and (4) by sampling solutions of $F$ using `SampleSat`, $*$-propagating them to covers, and computing an estimate of the positive and negative marginals from these covers (the cover marginals). Note that in (4), we are sampling true covers and obtaining an estimate. An alternative approach is to use SP itself on $F$ to try to sample covers of $F$, but the issue here is that the problem of finding (non-trivial) covers is not self-reducible to the decision problem of whether covers exist, as shown in Section 2. Therefore, it is not clear whether SP can be used to actually find a cover, despite it approximating the cover marginals very well.

Recall that the SP-based decimation process works by identifying variables with *extreme magnetization*, fixing them, and iterating. We will therefore be interested mostly in what happens in the extreme magnetization regions in these plots, namely, the lower left corner $(-1, -1)$ and the upper right corner $(+1, +1)$.

In the left pane of Figure 4 we plot the magnetization computed by SP on the x-axis and the magnetization obtained from cover marginals on the y-axis. The scatter plot has exactly 5000 data points, with one point for each variable of the formula $F$. If the magnetizations on the two axes matched perfectly, all points would fall on a single diagonal line from the bottom-left corner to the top-right corner. The plot shows that *SP is highly accurate at computing cover marginals, especially in the extreme regions* at the bottom-left and top-right.

The middle pane of Figure 4 compares the magnetization based on cover marginals with the magnetization based on solutions marginals. This will provide an intuition for why it might be better to follow cover marginals rather than solution marginals when looking for a satisfying assignment.[4] We see an interesting "s-shape" in this plot, which can be interpreted as follows: fixing variables with extreme cover magnetizations is more conservative compared to fixing variables with extreme solution magnetizations. Which means that variables that are extreme w.r.t. cover-based magnetization are also extreme w.r.t. solution-based magnetization (but not necessarily vice-versa). Recall that the extreme region is exactly where decimation-based algorithms, that often fix a small set of extreme variables per iteration, need to be correct. Thus, etimates of cover marginals provide a safer heuristic for fixing variables than estimates of solution marginals.

[3] For clarity, the plots show magnetizations for one such formula, although the trend is generic.

[4] Of course, if solution marginals could be computed perfectly, this would not be an issue. In practice, however, the best we can hope is to approximately estimate marginals.

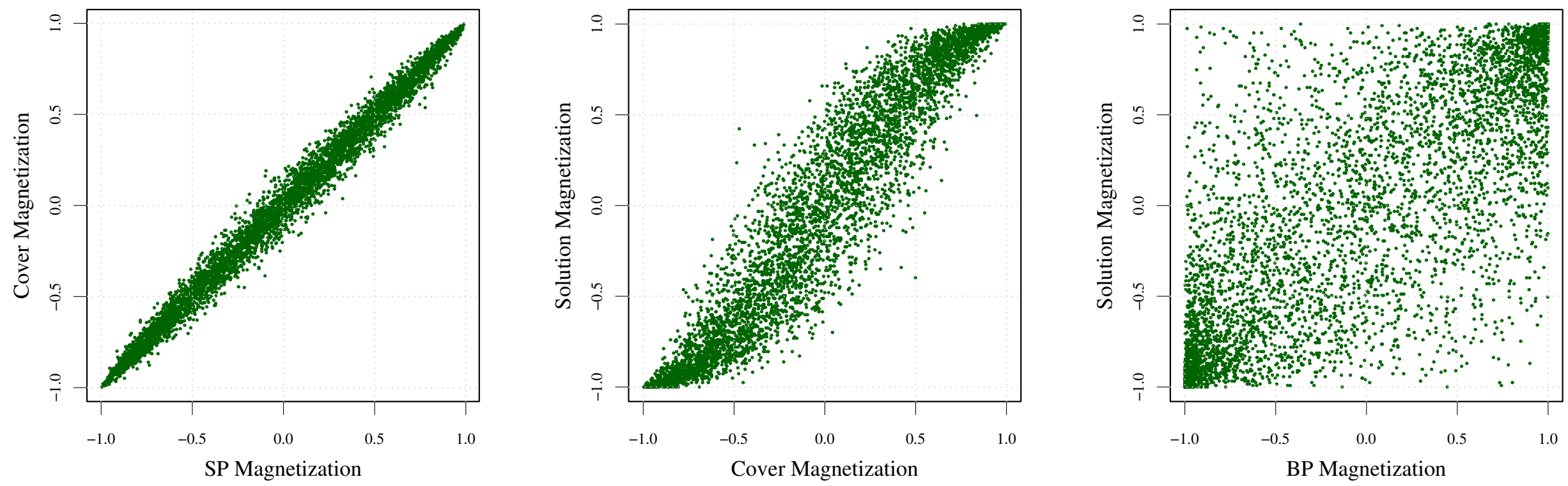

Figure 4: Magnetization plots. Left: SP vs. covers. Middle: covers vs. solutions. Right: BP vs. solutions.

As a comparison with BP, the right pane of Figure 4 shows BP magnetization vs. magnetization based on solution marginals for the same 5000 variable, 21000 clause formula. Since BP almost never converges on such formulas, we terminated BP after 10,000 iterations (SP took roughly 50 iterations to converge) and used the partially converged marginals obtained so far for computing magnetization. The plot shows that BP provides very poor estimates for the magnetizations based on solution marginals. (The points are equally scattered when BP magnetization is plotted against cover magnetization.) In fact, BP appears to identify as extreme many variables that have the opposite solution magnetization. Thus, when magnetization obtained from BP is used as a heuristic for identifying variables to fix, mistakes are often made that eventually lead to a contradiction, i.e. unsatisfiable reduced formula.

## 5 DISCUSSION

A comparison between left and right panes of Figure 4 suggests that approximating statistics over covers (as done by SP) is much more accurate than approximating statistics over solutions (as done by BP). This appears to be because covers are much more coarse grained than solutions; indeed, even an exponentially large cluster of solutions will have only a single cover as its representative. This cover still captures critical properties of the cluster necessary for finding solutions, such as backbone variables, which is what SP appears to exploit.

We also saw that the extreme magnetization based on cover marginals is more conservative than that based on solution marginals (as seen in the "s-shape" of the plot in the middle pane of Figure 4). This suggests that while SP, based on approximating cover marginals, may miss some variables with extreme magnetization, when it does find a variable to have extreme magnetization, it is quite likely to be correct. This provides an intuitive explanation of why the decimation process based on extreme SP magnetization succeeds with high probability on random 3-SAT problems without having to backtrack, while the decimation process based on BP magnetizations more often fails to find a satisfying assignment in practice.

We also note that BP and SP have been proven to compute exact marginals on solutions and covers, respectively, only for tree-structured formulas (with some simple exceptional cases like formulas with a single cycle). For BP, solution marginals on tree formulas are already non-trivial, and it is reasonable to expect it to compute a fair approximation of marginals on loopy networks (formulas). However, for SP, cover marginals on tree formulas are trivial: the only cover here is the all-$*$ cover. Cover marginals become interesting only when one goes to loopy formulas, such as random 3-SAT. In this case, as seen in our experiments, *it is remarkable that the SP computes a good approximation of non-trivial cover marginals for non-tree formulas.*

We hope that our results have convincingly demonstrated that the study of the covers of formulas is very fruitful and may well lead to a correct explanation of the success of SP.

## APPENDIX: DERIVATION OF THE SP EQUATIONS

Section 3 shows how to formulate a constraint satisfaction problem $P(F)$ such that its solutions are exactly the covers of a formula $F$. Here we proceed to show how the belief propagation formalism applied to $P(F)$ (as described in Section 3.1) results in the survey propagation equations.

**Review of BP.** We assume familiarity with the general form of BP equations, as used for example by Neapolitan (2004) in Theorem 3.2. In short, BP uses *messages* to communicate information between nodes of the factor graph (between variable nodes $x_a, \ldots$ and function nodes $F_x, F_a, \ldots$). Each message is a function of one argument, which takes on the same values as the variable node end-point of the message. There are two kinds of messages: from variable nodes to function nodes (denoted by $\pi_{x\to F}(.)$), and from function nodes to variable nodes (denoted by $\lambda_{F\to x}(.)$). In a two-level Bayesian Network, $\pi$ messages are computed by (piecewise) multiplying together the $\lambda$ messages received on all other links. The $\lambda$ messages are more complicated: they are sums across all possible worlds (values for arguments of received $\pi$ messages) of products of all-but-one $\pi$ messages with the chosen arguments. In case of a deterministic system (which is our case: every world has probability of either 1 or 0), this is equivalent to sum of products of $\pi$ messages with arguments that are compatible with each other as judged by the corresponding function node, $F_x$ or $F_a$. Moreover, if a variable node only has two neighboring function nodes, then it merely passes received $\lambda$ messages from one neighbor to the other. Since all variable nodes in $P(F)$ have degree two, we can safely ignore the existence of $\pi$ messages and only focus on $\lambda$ messages. Thus, every $F_x$ node receives messages from $F_a$ nodes (which we will denote by $\lambda_{a\to x}$) and and every $F_a$ node receives messages from $F_x$ nodes (denoted by $\lambda_{x\to a}$), both of which are functions of one argument, $(r_{a\to x}, w_{x\to a}) \in \{(0,0),(0,1),(1,0)\}$. The BP equations are constructed by considering the set of compatible variable node values given the one fixed value in the argument.

Let $C(x)$ be the set of all clauses containing variable $x$, and $V(a)$ the set of all variables appearing in clause $a$. Further, let $C_a^s(x)$ be the set of all clauses other than $a$ where $x$ occurs with the same sign as in $a$. Similarly define $C_a^u(x)$ to be the set of clauses where $x$ occurs with the opposite sign as in $a$. Note that $C_a^s(x) \cup C_a^u(x) \cup \{a\} = C(x)$.

**Equations for $F_x$.** The equations for messages sent from a factor node $F_x$ are given in Figure 5. For the argument value of $(1,0)$ the set of compatible values, as judged by $F_x$ when $x_a = (1,0)$, is one where there can be requests from clauses where $x$ appears with the same sign as in $a$ (and no warnings sent to them), but there must be no requests from opposite clauses (and warning must be sent). Similarly for $(0,1)$, but here the roles of $C_a^s(x)$ and $C_a^u(x)$ are exchanged, plus the fact that $x$ sends a warning to $a$ means that it must be receiving a request from some opposite clause (which is accounted for by the "−" term). Finally, for the value $(0,0)$, there are two possibilities: either at least one request is received from $C_a^s(x)$ (the first term in the sum, analogous to the expression for $(0,1)$), or there are no requests at all (the second term in the sum).

**Equations for $F_a$.** Figure 6 shows equations for messages sent from a factor node $F_a$. The argument value of $(1,0)$ is the easiest: $a$ sends out a request if and only if all other variables send it a warning, so that the only compatible values are all $(0,1)$. The case of $(0,0)$ is the complement: $a$ cannot receive a warning

$$\lambda_{x\to a}(1,0) = \prod_{b\in C_a^s(x)} (\lambda_{b\to x}(0,0) + \lambda_{b\to x}(1,0)) \prod_{b\in C_a^u(x)} \lambda_{b\to x}(0,1)$$

$$\lambda_{x\to a}(0,1) = \prod_{b\in C_a^s(x)} \lambda_{b\to x}(0,1) \left[\prod_{b\in C_a^u(x)} (\lambda_{b\to x}(0,0) + \lambda_{b\to x}(1,0)) - \prod_{b\in C_a^u(x)} \lambda_{b\to x}(0,0)\right]$$

$$\lambda_{x\to a}(0,0) = \left[\prod_{b\in C_a^s(x)} (\lambda_{b\to x}(0,0) + \lambda_{b\to x}(1,0)) - \prod_{b\in C_a^s(x)} \lambda_{b\to x}(0,0)\right] \prod_{b\in C_a^u(x)} \lambda_{b\to x}(0,1) + \prod_{b\in C(x)\setminus a} \lambda_{b\to x}(0,0)$$

Figure 5: BP equations for $F_x$

$$\lambda_{a\to x}(1,0) = \prod_{y\in V(a)\setminus x} \lambda_{y\to a}(0,1)$$

$$\lambda_{a\to x}(0,0) = \prod_{y\in V(a)\setminus x} (\lambda_{y\to a}(0,0) + \lambda_{y\to a}(0,1)) - \prod_{y\in V(a)\setminus x} \lambda_{y\to a}(0,1)$$

$$\lambda_{a\to x}(0,1) = \prod_{y\in V(a)\setminus x} (\lambda_{y\to a}(0,0) + \lambda_{y\to a}(0,1)) - \prod_{y\in V(a)\setminus x} \lambda_{y\to a}(0,1) + \sum_{y\in V(a)\setminus x} (\lambda_{y\to a}(1,0) - \lambda_{y\to a}(0,0)) \prod_{y'\in V(a)\setminus \{x,y\}} \lambda_{y'\to a}(0,1)$$

Figure 6: BP equations for $F_a$

from all other variables, and since $x$ does not send a warning, $a$ does not send a request anywhere. The last case, $(0,1)$, is a little more complicated. The first part is the same as before, but there needs to be a correction term to account for two extra possibilities: first it is now possible that $a$ issues a request to some $y$ if all other variables also send a warning (the positive term in the sum), and second it is not possible that all-but-one variable send $a$ a warning and yet $a$ does not issue a request to the last one (the negative term in the sum).

**Deriving the SP equations.** The expression for $\lambda_{a\to x}(0,1)$ can be simplified by assuming that $\lambda_{y\to a}(1,0) = \lambda_{y\to a}(0,0)$ for all $y$, in which case it reduces to the expression for $\lambda_{a\to x}(0,0)$. This assumption is crucial, but not very restrictive. Notice that it then follows that $\lambda_{x\to a}(1,0) = \lambda_{x\to a}(0,0)$ (by inspecting the appropriate expressions in Figure 5), and therefore the assumption keeps holding when iteratively solving the BP equations, provided it was true at the beginning.

The last step in the derivation is to rename and normalize the terms appropriately so as to "recognize" the SP equations in Figures 5 and 6. Let us define

$$\eta_{a\to x} \triangleq \prod_{y\in V(a)\setminus x} \frac{\lambda_{y\to a}(0,1)}{\lambda_{y\to a}(0,0) + \lambda_{y\to a}(0,1)}$$

and

$$\Pi^u_{x\to a} \triangleq \lambda_{x\to a}(0,1)$$
$$\Pi^0_{x\to a} \triangleq \prod_{b\in C(x)\setminus a} \lambda_{b\to x}(0,0)$$
$$\Pi^s_{x\to a} \triangleq \lambda_{x\to a}(0,0) - \Pi^0_{x\to a}$$

Notice that $\eta_{a\to x}$ is just a rescaled $\lambda_{a\to x}(1,0)$, and that the scaling factor $\lambda_{y\to a}(0,0) + \lambda_{y\to a}(0,1)$ equals $\Pi^u_{y\to a} + \Pi^s_{y\to a} + \Pi^0_{y\to a}$. Rescaling $\lambda_{a\to x}(0,0)$ and $\lambda_{a\to x}(0,1)$ in the same way (and using the assumption that they are equal) yields $\lambda_{a\to x}(0,0) = \lambda_{a\to x}(0,1) = 1 - \eta_{a\to x}$. Finally, writing down the BP equations for $\lambda_{x\to a}(0,1)$ and $\lambda_{x\to a}(0,0)$ in terms of these new variables results in the familiar SP equations established in Braunstein et al. (2005):

$$\eta_{a\to x} = \prod_{y\in V(a)\setminus x} \frac{\Pi^u_{y\to a}}{\Pi^u_{y\to a} + \Pi^s_{y\to a} + \Pi^0_{y\to a}}$$

$$\Pi^u_{x\to a} = \prod_{b\in C_a^s(x)} (1-\eta_{b\to x}) \left[1 - \prod_{b\in C_a^u(x)} (1-\eta_{b\to x})\right]$$

$$\Pi^s_{x\to a} = \prod_{b\in C_a^u(x)} (1-\eta_{b\to x}) \left[1 - \prod_{b\in C_a^s(x)} (1-\eta_{b\to x})\right]$$

$$\Pi^0_{x\to a} = \prod_{b\in C(x)\setminus a} (1-\eta_{b\to x})$$

In addition, the expressions for marginal probabilities computed by BP from a fixed point of the above equations can be shown, in a similar way, to be equivalent to the SP "bias" expressions.